\documentclass[letterpaper, 10 pt, conference]{ieeeconf}
\IEEEoverridecommandlockouts
\makeatletter
\let\NAT@parse\undefined
\makeatother

\usepackage{graphicx}
\usepackage{amsmath, amssymb, amsfonts}
\usepackage{hyperref}
\usepackage{nicematrix}

\title{\LARGE \bf
Learning Dexterous Bimanual Catch Skills through Adversarial-Cooperative Heterogeneous-Agent Reinforcement Learning
}
\author{Taewoo Kim, Youngwoo Yoon, and Jaehong Kim
\thanks{Taewoo Kim, as a Senior Researcher, is with the Social Robotics Research Section, Electronics and Telecommunications Research Institute (ETRI), Daejeon, Republic of Korea (e-mail: {\tt\small twkim0812@etri.re.kr}; {\tt\small twkim0812@gmail.com})}%
\thanks{Youngwoo Yoon, as a Principal Researcher, is with the Social Robotics Research Section, Electronics and Telecommunications Research Institute (ETRI), Daejeon, Republic of Korea (e-mail: {\tt\small youngwoo@etri.re.kr})}%
\thanks{Jaehong Kim, as a Principal Researcher and Director, is with the Social Robotics Research Section, Electronics and Telecommunications Research Institute (ETRI), Daejeon, Republic of Korea (e-mail: {\tt\small jhkim504@etri.re.kr})}%
}

\begin{document}

\maketitle
\thispagestyle{empty}
\pagestyle{empty}

%%%%%%%%%%%%%%%%%%%%%%%%%%%%%%%%%%%%%%%%%%%%%%%%%%%%%%%%%%%%%%%%%%%%%%%%%%%%%%%%
\begin{abstract}
%In this paper, we present a novel approach to learning bimanual catch skills through the adversarial cooperation of throw and catch agents, using Heterogeneous-Agent Reinforcement Learning (HARL). Our approach introduces an adversarial curricular reward structure where the throw agent increases the challenge for the catch agent by throwing objects with higher speed and unpredictability, while the catch agent improves its ability to catch under extreme conditions. The proposed method is validated in simulated environments with 15 different objects, showcasing its versatility and robustness across various scenarios. Key contributions include (1) exploring bimanual catching of diverse objects, (2) employing HARL to design an adversarial cooperation framework, and (3) demonstrating the effectiveness of our approach through extensive simulation experiments.

Robotic catching has traditionally focused on single-handed systems, which are limited in their ability to handle larger or more complex objects. In contrast, bimanual catching offers significant potential for improved dexterity and object handling but introduces new challenges in coordination and control. In this paper, we propose a novel framework for learning dexterous bimanual catching skills using Heterogeneous-Agent Reinforcement Learning (HARL). Our approach introduces an adversarial reward scheme, where a throw agent increases the difficulty of throws—adjusting speed—while a catch agent learns to coordinate both hands to catch objects under these evolving conditions. We evaluate the framework in simulated environments using 15 different objects, demonstrating robustness and versatility in handling diverse objects. Our method achieved approximately a 2x increase in catching reward compared to single-agent baselines across 15 diverse objects.

\end{abstract}

%%%%%%%%%%%%%%%%%%%%%%%%%%%%%%%%%%%%%%%%%%%%%%%%%%%%%%%%%%%%%%%%%%%%%%%%%%%%%%%%
\section{Introduction \& Related Work}

%In human environments, tasks like catching objects demand rapid decision-making and precise motor coordination. Bimanual catching, involving both hands, offers greater stability and control compared to one-handed methods, particularly when handling larger or more fragile objects. For robots to seamlessly collaborate with humans, mastering such dexterous skills is essential. However, most research in robotics manipulation has focused on slower, more controlled tasks, such as pick-and-place operations, where imitation learning techniques are effective in predictable environments. These methods, while successful in certain contexts, struggle to handle high-speed, dynamic tasks like catching. 

%Previous robotic studies in object catching have largely focused on using a single arm for both throwing and catching\cite{jaderberg2017population}, typically employing reinforcement learning (RL) to optimize performance. Although such approaches yield results, they overlook the benefits of bimanual coordination, which enhances task robustness and stability. Despite the advantages, the study of bimanual catching remains relatively underexplored in the field.

For robots to collaborate effectively with humans and respond swiftly to hazardous situations, such as catching falling objects, the ability to develop precise object-catching skills is crucial. Object catching is inherently challenging, even for humans, as it demands rapid decision-making, precise motor control, and real-time adaptation to dynamic environments. Several attempts have been made to imbue robots with dexterous skills, including object catching, though these efforts have predominantly focused on single-arm coordination.

For instance, DexPBT \cite{petrenko2023dexpbt} leveraged Population Based Training \cite{jaderberg2017population} to enable robots to learn complex dexterous tasks like throwing and object reorientation. Similarly, Dynamic Handover \cite{huang2023dynamic} simultaneously trained both a throwing robot and a catching robot to perform coordinated object manipulation tasks. Moreover, DexCatch \cite{lan2023dexcatch} advanced robotic catching skills by incorporating more realistic objects, such as cups and bananas, into the learning process.

Despite these advancements, a notable limitation across all these works is the reliance on single-arm catching. The potential of bimanual coordination has been largely overlooked, even though it offers clear advantages such as increased robustness, stability, and the ability to handle larger or irregularly shaped objects. The introduction of bimanual systems, as demonstrated by Bi-DexHands \cite{chen2023bi}, provided a manipulation framework and test environment for dual-arm tasks, but even this work restricted its catching operations to a single hand.

In parallel, multi-agent reinforcement learning (MARL) has emerged as a powerful framework for training agents in environments where multiple entities interact and learn simultaneously \cite{busoniu2008comprehensive, bucsoniu2010multi, zhang2021multi, hu1998multiagent}. MARL enables agents to learn optimal policies not only based on their interactions with the environment but also considering the behaviors of other agents, making it well-suited for complex coordination and competition tasks. Applications of MARL span various domains, including robotic swarm coordination \cite{yang2018mean}, autonomous driving \cite{shalev2016safe, zhang2024multi}, and collaborative manipulation \cite{gupta2017cooperative, de2020deep}.
%Within MARL, Heterogeneous-Agent Reinforcement Learning (HARL) focuses on scenarios where agents have different capabilities, action spaces, or objectives \cite{zhong2021heterogeneous}. This is particularly relevant for tasks involving agents with distinct roles, such as a thrower and a catcher in an object-catching scenario. HARL allows for the modeling of complex interactions between different types of agents, facilitating both cooperative and competitive behaviors. Previous works have leveraged HARL for tasks like heterogeneous robot teams performing search and rescue \cite{huttenrauch2019deep} and mixed-autonomy traffic control \cite{vinitsky2018benchmarks}.
However, the application of MARL to bimanual robotic catching remains underexplored. Most existing MARL research in robotics has focused on scenarios involving multiple robots or agents working independently or in loosely coupled ways. The integration of MARL techniques into the domain of bimanual manipulation presents an opportunity to enhance coordination and learning efficiency between robotic limbs.

\begin{figure}
    \begin{center}
    \includegraphics[width=\linewidth]{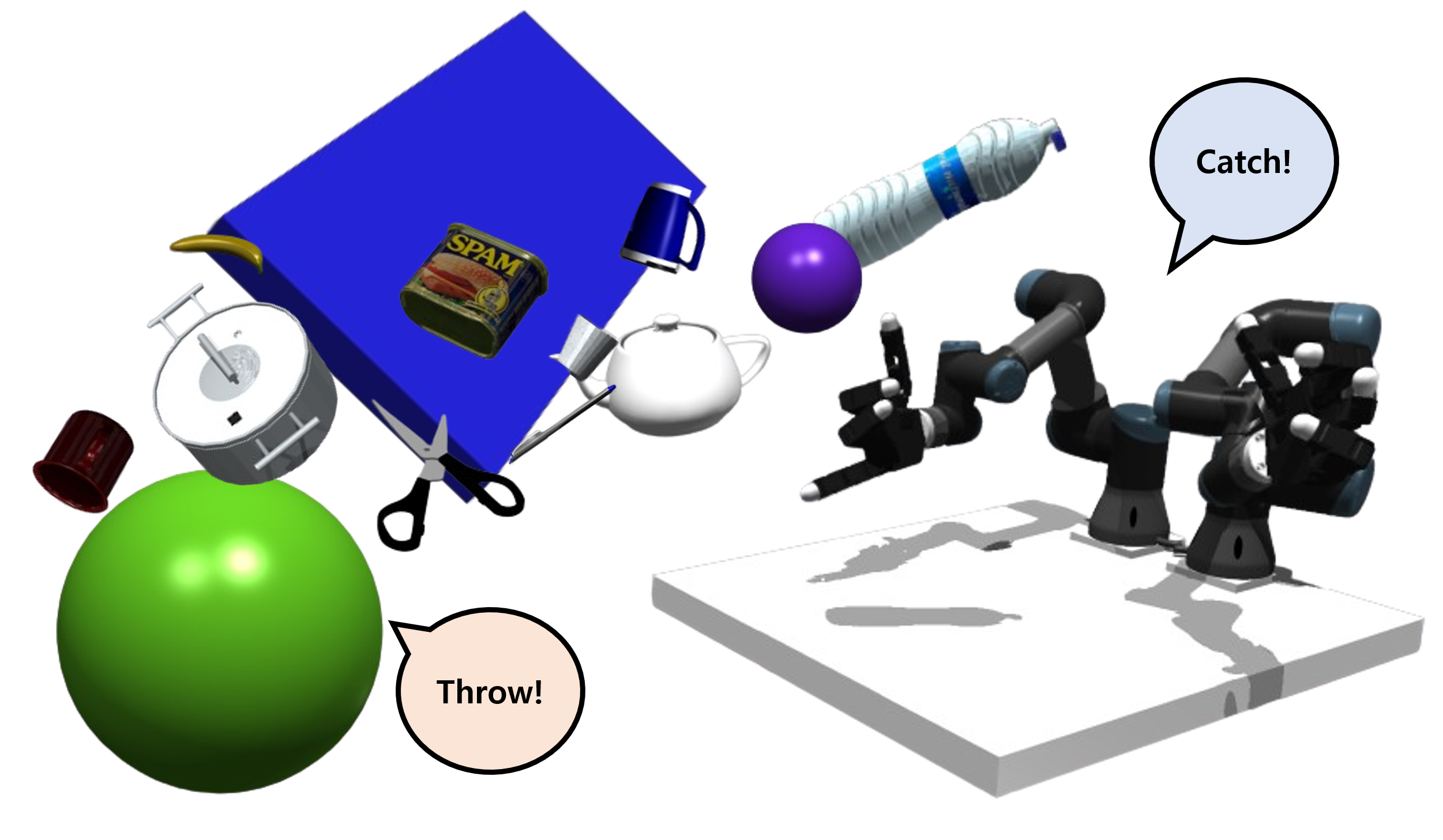}
    \end{center}
    \vspace{-1.5em}
    \caption{Dexterous bimanual catch skill is learned through adversarial cooperation. The catch agent with two arms and hands and the virtual throw agent are modeled for heterogeneous-agent reinforcement learning.}
    \label{fig:fig1}
    \vspace{-1.5em}
\end{figure}

% adversarial-cooperative 부연 설명
% 

\begin{figure*}[h!]
    \begin{center}
    \includegraphics[width=1.0\linewidth]{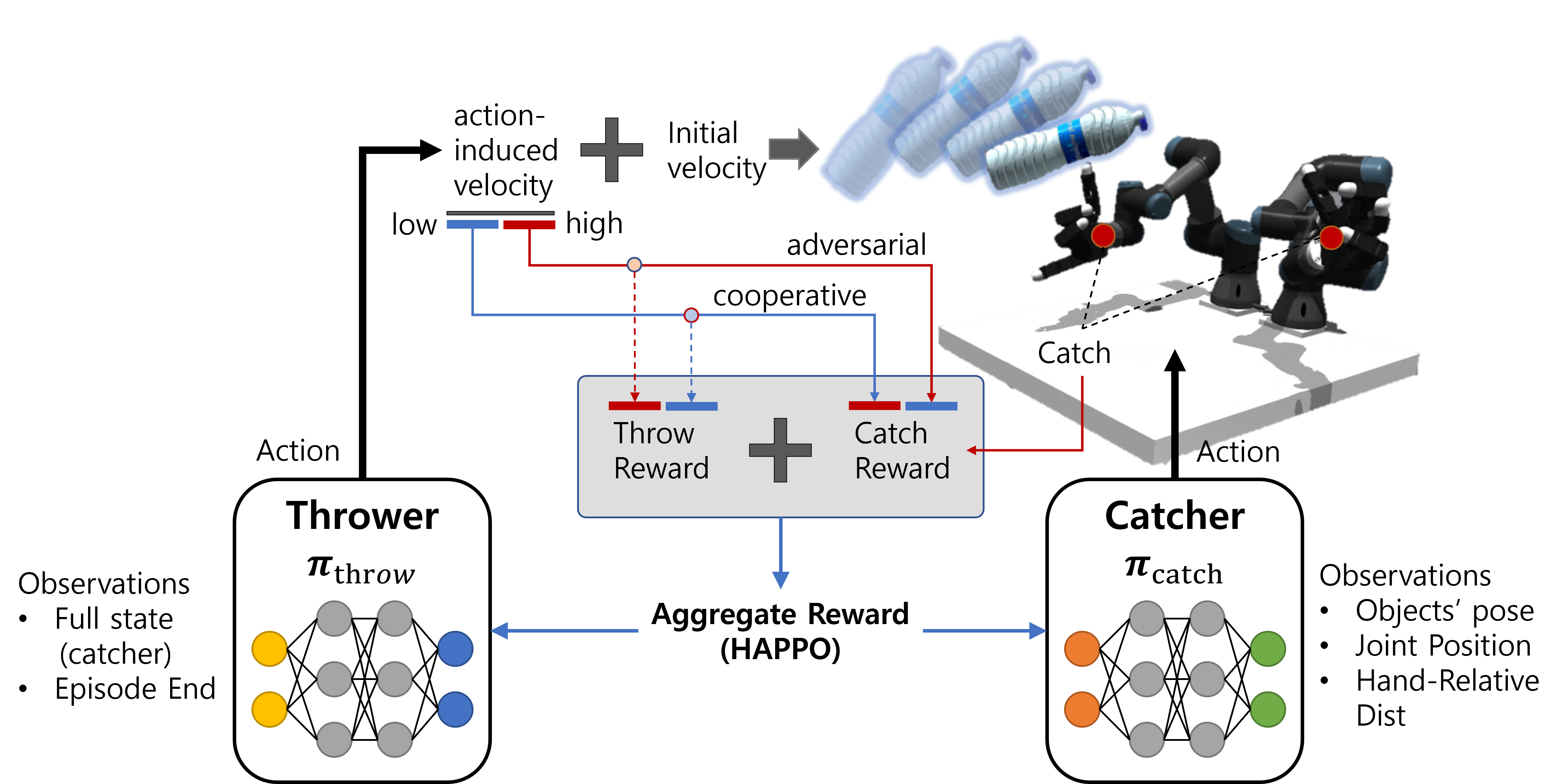}
    \end{center}
    \vspace{-1.0em}
    \caption{Throw-catch agents in a heterogeneous-agent RL framework. The throw agent, $\pi_{\text{throw}}$, sets the initial velocity of a random object at the start of each episode, while the catch agent, $\pi_{\text{catch}}$, learns to catch using a reward based on the object's distance to both hands. The adversarial-cooperative reward structure, shown in the center, reflects a paradoxical balance: when the action-induced velocity is high (red), the catch reward decreases due to adversarial difficulty, but the thrower gains a high reward. Conversely, when the velocity is low (blue), the catch reward increases, but the thrower receives a lower reward, highlighting the cooperative aspect. The aggregate reward is used for training via the HAPPO algorithm.}
    \label{fig:method}
    \vspace{-1.0em}
\end{figure*}

In this work, we propose a novel approach for bimanual catching skill development using Heterogeneous-Agent Reinforcement Learning. We structure the task as a collaborative environment with two agents: a thrower and a catcher, optimized through an adversarial reward system. By modeling the bimanual catcher as a single agent and introducing a virtual thrower agent, we harness the strengths of HARL to facilitate complex interaction dynamics. This approach enables the development of robust and dexterous catching capabilities, even under extreme conditions, offering significant advancements in robotic manipulation.

We present three key contributions in this paper:

%1) \textbf{Tackling Multi-Object Bimanual Catching.} We address the relatively unexplored area of bimanual catch learning from the perspective of multi-object catch tasks, moving beyond the single-object, one-handed exchange studies prevalent in prior research.

% Multi-Object 가 오해의 소지가 있어서 아래와 같이 변경 + 문구 약간 수정
1) \textbf{Tackling Bimanual Catching Skill.} We explore the relatively under-explored domain of bimanual catching skill learning in robots. %, handling a diverse range of objects. 
This research advances beyond the common one-handed catching approaches %with simple objects 
in previous work.

%2) \textbf{Adversarial Curriculum Learning.} We propose an adversarial curriculum learning method based on HARL, enabling robots to handle extreme scenarios requiring heightened dexterity and robustness. By modeling the thrower and catcher as heterogeneous agents with competing objectives, we facilitate a learning environment where both agents improve through continuous adaptation.

2) \textbf{Adversarial-Cooperative Learning Framework.} We introduce an adversarial-cooperative reward structure within the HARL framework, enabling robots to learn robust bimanual catching skills under challenging and dynamic conditions. By modeling the thrower and catcher as heterogeneous agents with both adversarial and cooperative objectives, we create a learning environment where continuous adaptation leads to mutual improvement.

%3) \textbf{Validation via Simulation.} Through extensive simulation experiments, we demonstrate the effectiveness of our approach across various object sizes and types, highlighting the versatility of the proposed bimanual catch framework.

3) \textbf{Validation with Diverse Objects.} Through extensive simulation experiments, we demonstrate the effectiveness of our approach across various object sizes and types, highlighting the versatility of the proposed bimanual catch framework.

The code is available at \href{https://github.com/taewookim0812/BimanualDexCatch}{this github}.

\section{Throw-Catch Agents for Learning Bimanual Catch Skills}

% our bimanual catcher agent는 두 개의 UR3와 left and right allegro hand로 구성되며, workspace table에 고정되어 있다. 
% catcher의 observation은 proprioceptive 정보인 6자유도 arm 관절, 16자유도의 핸드 joint정보, 그리고 양손의 손바닥 중심부에 대한 pose(7-dim)정보의 pair로 구성되며, 15개 모든 object의 pose(7-dim) 및 양손과의 상대 벡터 정보를 함께 포함하여 총 253차원으로 구성된다. catcher의 action은 두 arm과 양손에 대한 joint torque 명령어로 정의되며, 총 44차원의 action으로 정의된다 (see Figure 2). 

\subsection{Bimanual Catcher Agent}

Our bimanual catcher agent is composed of two UR3 robotic arms \cite{UR3}, each equipped with left and right Allegro hands \cite{AllegroHand_TechInfo}, both securely mounted on the workspace table. The catcher’s observation space encompasses proprioceptive data from the 6-DoF arm joints and 16-DoF hand joints, as well as paired pose information for the center of each palm. It also includes 7-dimensional poses for each object (3 for position and 4 for quaternion rotation) as well as 3-dimensional relative vectors between the object and each palm center, for all 15 objects. In total, this results in an observation space of 253 dimensions. The agent’s action space comprises joint torque commands for the arms and hands, forming a 44-dimensional space (see Figure \ref{fig:method}).

% The agent’s action space consists of joint torque commands for both the arms and hands, yielding a 44-dimensional action space (see Figure \ref{fig:method}).

% thrower는 환경에서 직접적으로 보이지는 않으나, 던지기 위해 선정된 object에 linear, angular velocity를 부여하는데 관여한다. 
% 매 에피소드의 시작 시점에 임의로 선정된 object는 던지기 starting point에 uniform noise를 더해 소환하고, 마찬가지로 uniform noise가 가미된 catcher방향으로의 throwing을 위한 기본 속도가 할당된다. Single-agent기반 학습은 이와 같은 기본 절차를 통해 object를 자동으로 던지며 학습이 진행되나, 본 연구에서 제안하는 HARL 프레임워크에서는 선정된 object의 초기 속도에 thrower의 action이 더해져서 다음 식과 같이 object의 최종 초기 속도가 결정된다:

\subsection{Thrower Agent}

The thrower agent, although not physically represented in the environment, is crucial for determining the initial conditions of the thrown objects. In the default policy used for single-agent learning, objects are initialized with random positions and assigned base velocities towards the catcher, both perturbed by uniform noise. However, in our HARL framework, the thrower agent learns to adjust these velocities to challenge the catcher. Specifically, the thrower policy $\pi_{\theta_{\text{throw}}}$ outputs an action $\boldsymbol{v}_{\text{action}}$ that modifies the base velocity: 
\begin{align} 
    \boldsymbol{v} &= (\boldsymbol{v}_{\text{base}} \times \epsilon) + \boldsymbol{v}_{\text{action}},
\end{align} 
where $\boldsymbol{v} = \{ v_{\text{x}}, v_{\text{y}}, v_{\text{z}}, v_{\text{rx}}, v_{\text{ry}}, v_{\text{rz}}\}$ represents the final linear and angular velocities, and $\epsilon \sim \mathcal{U}(-0.5, 0.5)$ is uniform noise. The addition of $\boldsymbol{v}_{\text{action}}$ allows the thrower to influence the difficulty of the catch task.

This design choice ensures stable learning by preventing the thrower from generating erratic or overly challenging throws that could disrupt the catcher's learning. Anchoring the thrower's actions to a reasonable base velocity creates a balanced environment, enabling both agents to improve their policies without causing instability or convergence issues.

In conventional multi-agent RL \cite{busoniu2008comprehensive}, agents output actions at each step based on their observations, influencing the environment by moving their respective components. 
%However, in this study, the thrower agent’s action affects the object’s velocity only in the first frame of each episode, after which the object’s state is determined by the physics engine, rendering the thrower’s actions for the remaining frames irrelevant. 
However, in this study, the thrower agent’s effective action is throwing velocity of objects which happens only in the first frame of each episode, after which the object’s state is determined by the physics engine. 
To provide a cue regarding the episode’s initiation, the thrower’s observation is augmented with a 1-dimensional episode termination signal added to the catcher’s observation, resulting in a total of 254 dimensions. The thrower’s observation includes the state of all objects and the catcher to observe how its actions affect the environment, enabling more stable policy learning within this simulation environment. The thrower’s action space is composed of 6 dimensions, representing the linear $(\text{meter}/\text{second})$ and angular $(\text{radian}/\text{second})$ velocities applied to the object.

% 기본적으로 multi-agent RL에서는 모든 에이전트들이 매 step마다 주어진 observation으로부터 action을 출력하고 이를 통해 자신의 본체를 움직여서 환경에 영향을 주게 된다. 그러나 본 연구에서 thrower agent의 action은 오직 에피소드의 첫 프레임에서만 object의 속도에 영향을 주고 이후 물리엔진에 의해 object의 상태가 결정되고, 나머지 프레임에서는 의미 없는 행동을 추론하게 된다. 따라서 에피소드의 시작 정보에 대한 힌트를 주기 위하여 thrower의 observation은 catch observation에 1-dimensional 에피소드 종료 정보를 포함하여 총 254-dimensional observation으로 구성되고, action은 linear 및 angular adding velocity의 6차원으로 구성된다. 

\subsection{Objects}

% 제안한 학습 프레임워크를 통해 bimanual robot agent가 다양한 종류의 물체를 잡을 수 있는지 시험하기 위하여, 타겟 object로 다양한 형태의 크고 작은 물체를 선정하였다. 한 손으로도 잡을 수 있는 크기가 작은 물체부터 반드시 두 손을 써야 잡을 수 있는 큰 물체까지 총 15종의 물체를 선정하였으며, primitive objects, Bi-dexhands objects, 그리고 YCB objects들을 포함한다: 

%To evaluate whether the proposed learning framework enables the bimanual robot agent to catch a wide variety of objects, we selected a diverse set of target objects, ranging from small items that can be caught with one hand to larger objects that require both hands. A total of 15 objects were chosen, including primitive objects \{Gymball, Bowling ball, Cube, Board\} and a subset of objects used in YCB \cite{ycb} \{Banana, Meat can, Mug, Brick\} and Bi-DexHands \cite{chen2023bi} \{Kettle, Bottle, Cup, Bucket, Pen, Pot, Scissors\} (see Figure \ref{fig:objects}).
% \begin{itemize}
%     \item Primitive: \{Gymball, Bowling ball, Cube, Board\}
%     \item YCB: \{Banana, Meat can, Mug, Brick\}
%     \item Bi-DexHands: \{Kettle, Bottle, Cup, Bucket, Pen, Pot, Scissors\}
% \end{itemize}

\begin{figure}
    \begin{center}
    \includegraphics[width=\linewidth]{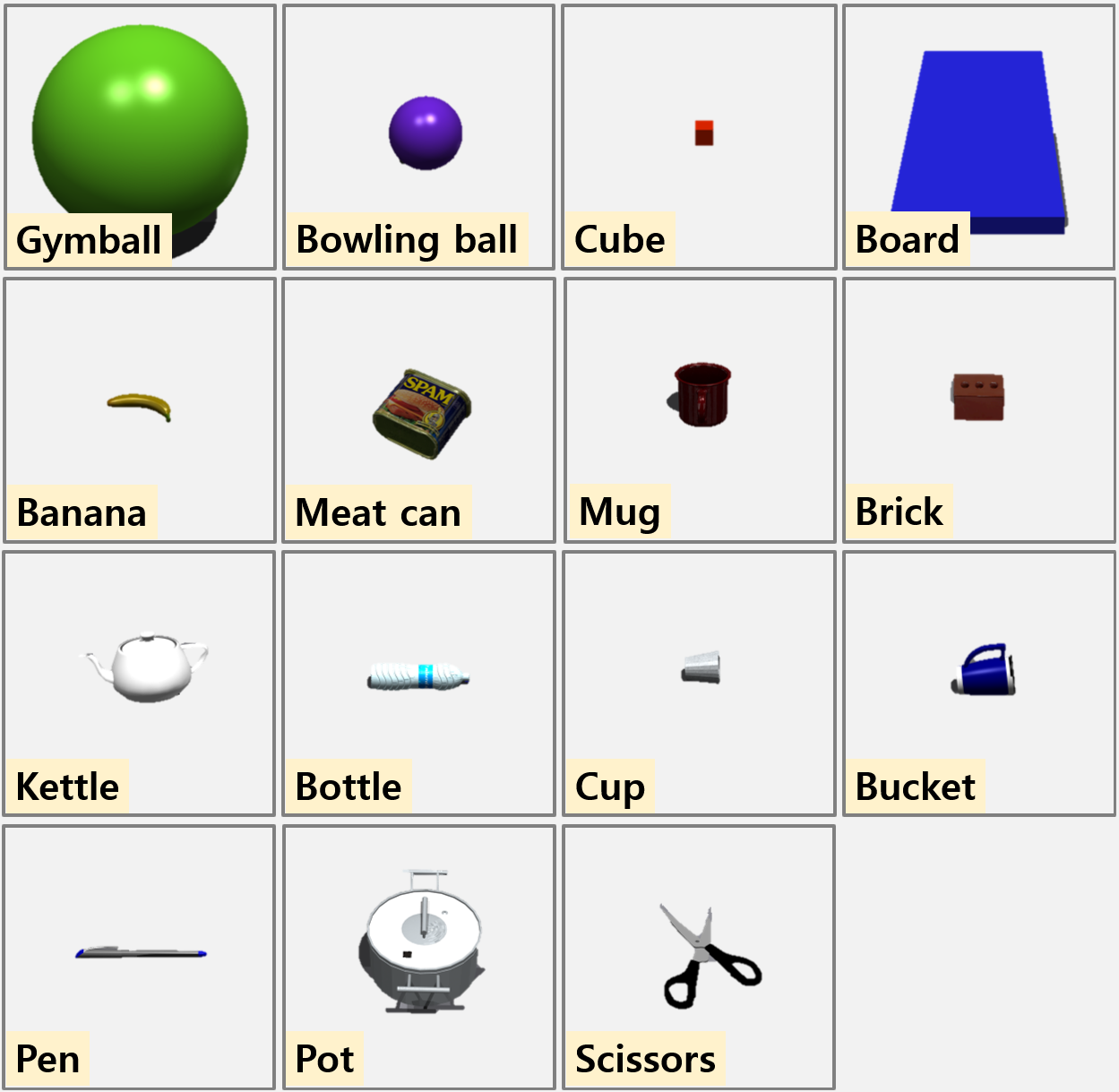}
    \vspace{-2.5em}
    \end{center}
    \caption{15 objects in different sizes and shapes used in the experiments.}
    \label{fig:objects}
    \vspace{-1.5em}
\end{figure}

To evaluate the framework’s effectiveness in bimanual catching, we selected 15 objects varying in size and handling requirements, from small, single-hand items to larger, two-hand objects. The selection includes primitive objects \{Gymball (0.6m radius), Bowling ball (0.215m radius), Cube (0.05m edge), Board (0.9m × 0.4m × 0.05m)\}, as well as subsets from YCB \cite{ycb} \{Banana, Meat can, Mug, Brick\} and Bi-DexHands \cite{chen2023bi} \{Kettle, Bottle, Cup, Bucket, Pen, Pot, Scissors\}, which were directly adopted from their respective datasets (see Figure \ref{fig:objects}).

%To assess the framework's ability for bimanual catching, we selected 15 objects, from small, single-hand to larger, two-hand objects. These include primitive objects \{Gymball, Bowling ball, Cube, Board\}, YCB \cite{ycb} \{Banana, Meat can, Mug, Brick\} and Bi-DexHands \cite{chen2023bi} \{Kettle, Bottle, Cup, Bucket, Pen, Pot, Scissors\} (see Figure \ref{fig:objects}).

\begin{figure*}
    \begin{center}
    \includegraphics[width=0.9\linewidth]{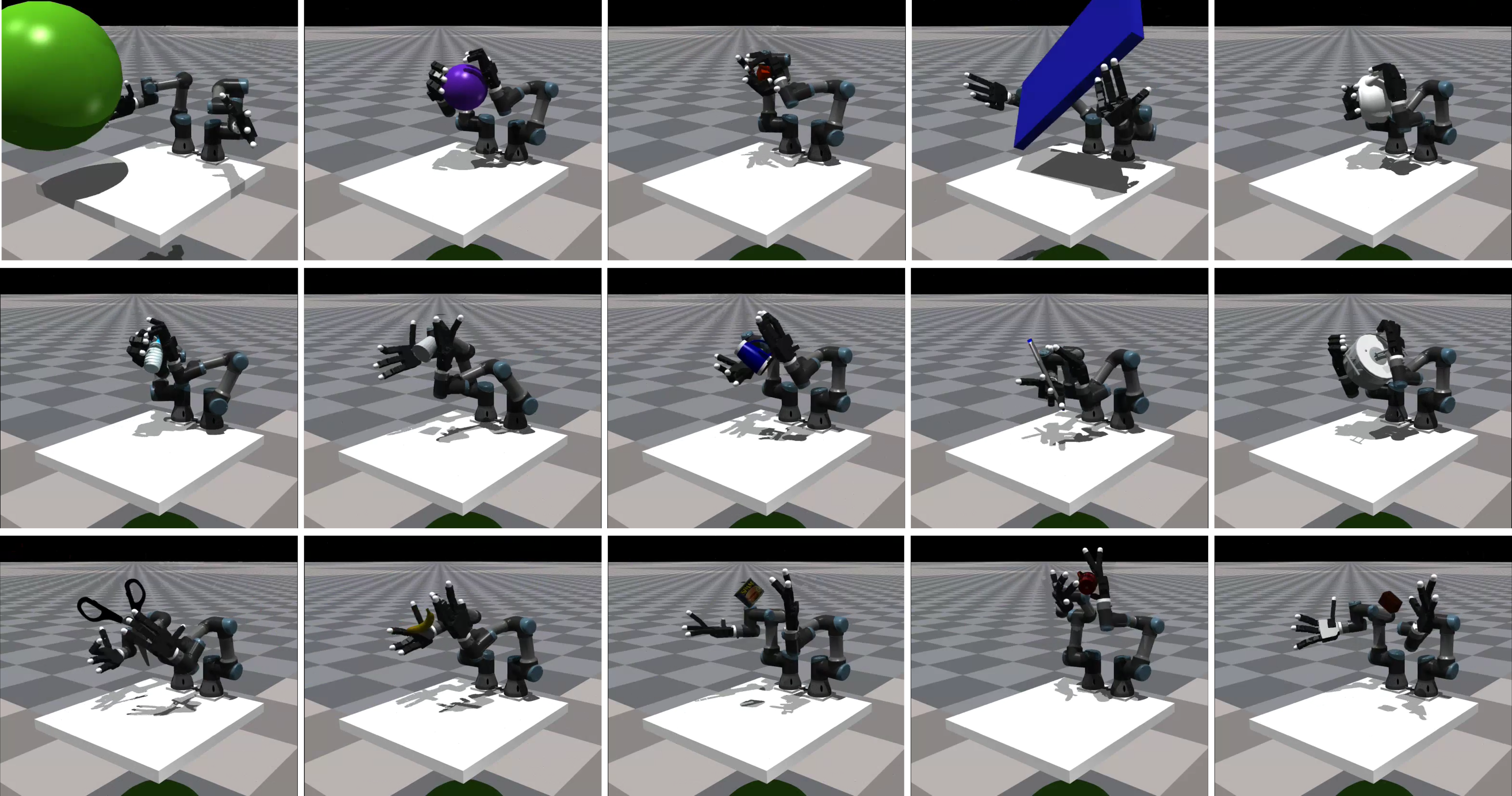}
    \end{center}
    \vspace{-1.5em}
    \caption{Catch agent performing bimanual catching of 15 different objects thrown randomly.}
    \label{fig:fig4}
    \vspace{-1.5em}
\end{figure*}

%%%%%%%%%%%%%%%%%%%%%%%%%%%%%%%%%%%%%%%%%%%%
% HARL 이해를 위한 부연 설명
% MAPPO와 HAPPO의 결정적인 차임
% MAPPO는 단순 parameter sharing을 통해 모든 에이전트를 중앙집중화된 체계하에서 학습
% 구체적으로는 모든 에이전트의 학습에 동일하게 사용되는 global critic network이 있어서 이 parameter가 공유되는 것임
% 반면 HAPPO는 개별 Critic 네트워크가 있어서 보다 독립적인 학습이 가능함. 
% 수식에서 compound policy ratio는 importance sampling(IS)을 모든 agent에 걸쳐서 global하게 하는 것을 의미함. IS는 이전 policy의 행동 확률과 현재 policy의 행동 확률 사이의 ratio에 비례하여 Advantage를 적용하여 좋은 행동을 학습하게 되는데, IS는 이전 확률과의 비율에 따라 너무 급격한 행동 변화를 방지하여 학습 안정성을 높이고 발산을 방지하는 것임. 이것이 TRPO의 trust-region을 정의하는데에 활용되는데, 
% HAPPO는 여러 agent에 걸쳐서 global하게 IS를 해서 전체적인 학습 안정성을 높이고, 이전 정책 활용에 따른 효율적 학습, 탐험-활용 균형, off-policy learning등 다양한 목적을 가짐
% 결론적으로 MAPPO와 HAPPO의 차이를 가르는 가장 중요한 요소는 parameter shared critic의 존재에 따라 전역성을 갖느냐, 혹은 보다 독립성을 갖느냐의 차이임. 또한 동일한 형태의 복수 agent냐(e.g., 캐리어의 인터셉터들), 혹은 완전히 독립적인 서로 다른 개체이냐(e.g., 캐리어와 배틀크루저)에 따라 그 개념이 나뉨. 

\section{Adversarial Cooperation between Throw-Catch Agents}

\subsection{Multi-Agent RL for Throw-Catch System} 
We model the system as a multi-agent framework using Heterogeneous-Agent Reinforcement Learning (HARL) \cite{zhong2024heterogeneous}. Adhering to the Centralized Training with Decentralized Execution (CTDE) paradigm in multi-agent reinforcement learning (MARL) \cite{busoniu2008comprehensive}, both the thrower and catcher agents optimize the following HAPPO objective function:
\begin{equation}
\begin{aligned}
    \mathbb{E}_{s \sim \rho_{\boldsymbol{\pi_{\theta_k}}}, \text{a} \sim \boldsymbol{\pi_{\theta_k}}} &\left[ \min \left( \frac{\pi^{i_m}_{\theta^{i_m}}(\text{a}^{i_m} | s)}{\pi^{i_m}_{\theta_k^{i_m}}(\text{a}^{i_m} | s)} M^{i_{1:m}}(s, \textbf{a}), \right. \right. \\
    &\quad \left. \left. \text{clip} \left( \frac{\pi^{i_m}_{\theta^{i_m}}(\text{a}^{i_m} | s)}{\pi^{i_m}_{\theta_k^{i_m}}(\text{a}^{i_m} | s)}, 1 \pm \epsilon \right) M^{i_{1:m}}(s, \textbf{a}) \right) \right]
\end{aligned}
\end{equation}

% \noindent where each agent $i_m$ optimizes the joint clipped surrogate objective for the multi-agent system, using the ratio of its current policy $\pi^{i_m}_{\theta^{i_m}}$ to its previous ($k$) policy $\pi^{i_m}_{\theta_k^{i_m}}$, represented in a generalized form, along with the compound policy ratio $M^{i_{1:m}}$, constrained by the PPO clipping range $1 \pm \epsilon$.

\noindent where each agent $i_m$ optimizes the joint clipped surrogate objective for the multi-agent system, using the ratio of its current policy $\pi^{i_m}_{\theta^{i_m}}$ to its previous policy $\pi^{i_m}_{\theta_k^{i_m}}$, along with the compound policy ratio $M^{i_{1:m}}$, constrained by the Proximal Policy Optimization (PPO) clipping range $1 \pm \epsilon$.
\begin{equation}
\begin{aligned}
    M^{i_{1:m}}(s, \textbf{a}) &= \frac{\pi_{\theta_{k+1}^{i_{1:m-1}}}(a^{i_{1:m-1}}|s)}{\pi_{\theta_k^{i_{1:m-1}}}(a^{i_{1:m-1}}|s)} \hat{A}(s, \textbf{a}) \\
\end{aligned}
\end{equation}

In this study, the compound policy ratio $M^{i_{1:m}}$ is computed jointly for the thrower agent policy ${\pi}_{\theta}^{i_\text{thrower}}$ and the catcher agent policy ${\pi}_{\theta}^{i_\text{catcher}}$ at each training iteration. The ratio is updated based on the relative changes between the previous policy parameters $\theta_k$ and the updated policy parameters $\theta_{k+1}$, applied in an arbitrary agent order \cite{zhong2024heterogeneous}.

\subsection{Adversarial-Cooperative Reward}

The thrower seeks to maximize its reward by generating fast throws, while the catcher aims to catch the objects successfully under diverse and extreme conditions. Both agents learns their strategies by optimizing their respective reward functions:
\begin{align}
    r_{\text{catch}} &= \begin{aligned}[t]
        &w_0 \, r_{\text{hand\_dist}} + w_1 \, r_{\text{goal}} + w_2 \, r_{\text{finger\_contact}} \\
        &- w_3 \, \bar{r}_{\text{arm\_contact}} - w_4 \, \bar{r}_{\text{catcher\_action}}
    \end{aligned} \\[0.5em]
    r_{\text{throw}} &= \begin{aligned}[t]
        &w_5 \, r_{\text{object\_velocity}} + w_6 \, r_{\text{thrower\_action}}
    \end{aligned}
\end{align}

\noindent where the catch reward is a weighted sum of sub-rewards for hand distance, goal proximity, and finger contact, with penalty terms for arm contact and action, represented by bar symbol respectively. $r_{\text{hand\_dist}}$ rewards the agent for minimizing the distance between the thrown object and the center of the palms, while $r_{\text{goal}}$ provides a higher reward as the object approaches the predefined goal position in front of the robot. $r_{\text{finger\_contact}}$ reward is introduced to promote grasping actions, particularly for smaller objects, by encouraging finger flexion, granting a reward when the fingertips make contact with objects. 
% (아래 문장은 어색해서 제외)
%Since the robot hands are the primary manipulators, 
A penalty term $\bar{r}_{\text{arm\_contact}}$
%is added to discourage collisions with the robot arms, guiding the catcher to use only its hands for grasping the object. This term 
is designed to minimize collisions between the left and right arms or between the arms and the object, encouraging the agent to avoid arm-object or arm-arm collisions during the catching process. Additionally, an action regularization term $\bar{r}_{\text{catcher\_action}}$ is included to minimize unnecessary motions and encourage efficient task execution. This term applies a small penalty proportional to the magnitude of the action values, a common reward shaping technique that discourages excessive movement while promoting controlled and effective catching. $r_{\text{throw}}$ is a combination of $r_{\text{object\_velocity}}$, which rewards higher linear and angular velocities of the thrown object, and $r_{\text{thrower\_action}}$, which provides a reward based on the thrower’s actions that induce these velocities. The weights $w_0$ to $w_6$ are set to \{5.0, 1.0, 0.5, 0.5, 1e-3, 0.8, 0.2\} respectively, determined through empirical tuning. Please refer to the code for exact implementation details.

% $r_{hand\_dist}$는 던져진 물체와 양손의 손바닥 중심부와의 거리가 가까울 수록 높은 보상을 얻고, $r_{goal}$은 물체가 로봇의 앞쪽에 설정된 goal position과 가까울 수록 높은 보상을 얻으며, $r_{lift}$는 특정 높이 이상 물체를 들어올린 경우에 보상을 얻도록 설계하였다. $r_{separate}$은 손끼리 충돌을 방지하고자 포함시켰으며, 양손의 거리가 멀 수록 높은 보상을 얻는다. $r_{finger\_contact}$은 비교적 작은 물체를 잡는 과정에서 손가락을 굽혀서 grasping을 하는 동작을 유도하기 위하여 설정하였으며, fingertip에 contact이 발생하면 보상을 얻도록 설계하였다. 물체를 잡는 주체는 로봇 핸드이기 때문에 가급적 로봇 arm과는 충돌이 발생하지 않도록 $\bar{r}_{\text{arm\_contact}}$ penalty를 부여하였으며, 최대한 적은 모션을 발생시켜서 작업을 수행할 수 있도록 action에 대한 penalty term $\bar{r}_{\text{action}}$도 추가하였다. 
% $r_{throw}$는 던지는 object의 linear 및 angular 속도를 빠르게 할 수록 높은 보상을 얻는 $r_{object\_velocity}$와 이를 유발하는 thrower의 행동에 따른 보상인 $r_{thrower\_action}$의 조합으로 구성된다. 
% 보상함수의 가중치 $w_0 \~ w_6$은 각각 \{5.0, 1.0, 0.5, 0.5, 1e-3, 0.8, 0.2\}으로 설정되었다. 

%In our framework, the interaction between the thrower and catcher agents naturally creates an implicit adversarial curriculum, governed by the following reward function:
Our framework combines $r_{\text{catch}}$ and $r_{\text{throw}}$ to use the following final reward:
\begin{align}
    r_{\text{total}} &= \begin{aligned}[t]
        &\alpha \, r_{\text{catch}} + (1-\alpha) \, r_{\text{throw}},
    \end{aligned}
\end{align}
%where $\alpha$ controls the balance between adversarial (lower $\alpha$) and cooperative (higher $\alpha$) learning dynamics.
where $\alpha$ controls the balance between adversarial (lower $\alpha$) and cooperative (higher $\alpha$) learning dynamics.
The thrower is rewarded for increasing the difficulty of the throw by maximizing the object’s speed and unpredictability, while the catcher is rewarded for successfully adapting to these challenging conditions. 

Although we do not explicitly define a curriculum for the thrower, the interaction between the thrower and catcher agents naturally would create an implicit adversarial curriculum to maximize the total reward. If the thrower throws the object too quickly, the catcher may fail to catch it, resulting in a lower catch reward. On the other hand, if the thrower throws too easily, the throw reward will decrease. This adversarial dynamic effectively simulates a curriculum, where the difficulty is automatically scaled based on the thrower’s evolving strategy. The catcher, in turn, develops more refined catching skills as it overcomes these challenges. Thus, the adversarial nature of the thrower-catcher relationship fosters an emergent cooperative effect, wherein the thrower indirectly guides the catcher to enhance its capabilities, resulting in a robust and adaptive learning process.

We determined $\alpha$ experimentally, and also tested a strategy of gradually decreasing $\alpha$ during the learning process, which progressively increases the throwing difficulty. Experimental results are presented in Section \ref{sec:evaluation_ma}. 
%the variable $\alpha$ allows us to adjust the degree to which the catcher's performance is prioritized over the thrower's adversarial behavior. As $\alpha$ decreases, the thrower exerts greater difficulty, forcing the catcher to adapt under increasingly extreme conditions. 

\subsection{Policy Network Architecture}

Both agents utilize multi-layer perceptrons (MLPs) with [1024, 512, 256] units for their policy networks, employing the exponential linear unit (ELU) activation function \cite{clevert2015fast}. Each agent follows an Actor-Critic architecture, trained with PPO \cite{schulman2017proximal} within the HARL framework \cite{zhong2024heterogeneous}. To improve the network's ability to handle complex, high-dimensional state spaces, we incorporate Deep Dense Architectures in Reinforcement Learning (D2RL) \cite{sinha2020d2rl}.

% 에피소드가 리셋 되면 새로운 object가 랜덤하게 선택되어 throw start position에 respawn되고, 기존에 던져졌던 object는 나머지 object들과 함께 staging area로 telepoert되어 에피소드가 끝날 때까지 대기한다. 

% \begin{figure}
% 	\begin{center}
% 		\includegraphics[width=\linewidth]{images/fig_staging.png}
% 	\end{center}
	
% 	\caption{At the start of each episode, a new object is randomly selected and respawned at the throw start position, while the previously thrown object is teleported to the staging area along with the remaining objects, where they remain until the episode ends.}
%         \label{fig:staging}
% 	\vspace{-1.5em}
% \end{figure}

\section{Experiments}

\subsection{Experimental Details}

The experiments were conducted on the IsaacGym simulation platform, with the code base built on the IsaacGymBenchmark environment \cite{issacgym}. 
We have implemented the HARL framework on top of \textit{rl\_games} library \cite{rl-games2021}. 
The bimanual robot system is positioned one meter above the ground plane alongside a workspace table. In the environment, 15 objects are initialized, from which the thrower randomly selects one to launch toward the catcher. At the beginning of each episode, the catcher resets to its initial position, while the previously thrown object is respawned in a staging area beneath the table. The thrower then selects another object to repeat the process.

The experiments were run on a workstation equipped with an AMD Ryzen 9 7950X 16-Core processor, 128GB RAM, and an RTX 6000 Ada GPU. Training for 10,000 epochs required approximately 8 hours with a batch size of 4,096 environments. To ensure diverse interactions, the object selected for throwing was uniformly reset to a random position at the beginning of each episode.

% IsaacSGym 시뮬레이션 플랫폼을 활용하여 개발, 우리의 기본 코드 베이스는 IsaacGymBenchmark환경이며, 이 환경의 rl_games library를 기반으로 HARL 프레임워크를 구현함. 우리의 bimanual robot 플랫폼은 작업 테이블과 함께 ground plane으로부터 1m높이의 공중에 떠 있음. 

% 15개의 object들은 환경에 모두 생성된 상태에서 작업 테이블 하단의 대기 장소에 respawn되고, thrower는 그 중 하나를 랜덤하게 골라서 catcher 방향으로 던진다. 던지고 받는 하나의 에피소드가 종료되면 catcher는 초기 자세로 리셋되며, 이전 에피소드에서 던졌던 object는 대기 장소에 소환되고 thrower는 다시 랜덤하게 object 하나를 골라서 catcher 방향으로 던지고 이 과정을 반복한다. 

% 학습에 활용한 mahcine은 AMD Ryzen 9 7950X 16-Core와 128GB RAM, RTX6000ada GPU를 가지고 있으며, 10K epoch 학습에 대략 8시간 이상 소요됨. 

\begin{figure}
	\begin{center}
		\includegraphics[width=\linewidth, trim=0cm 7.4cm 0cm 0cm, clip]{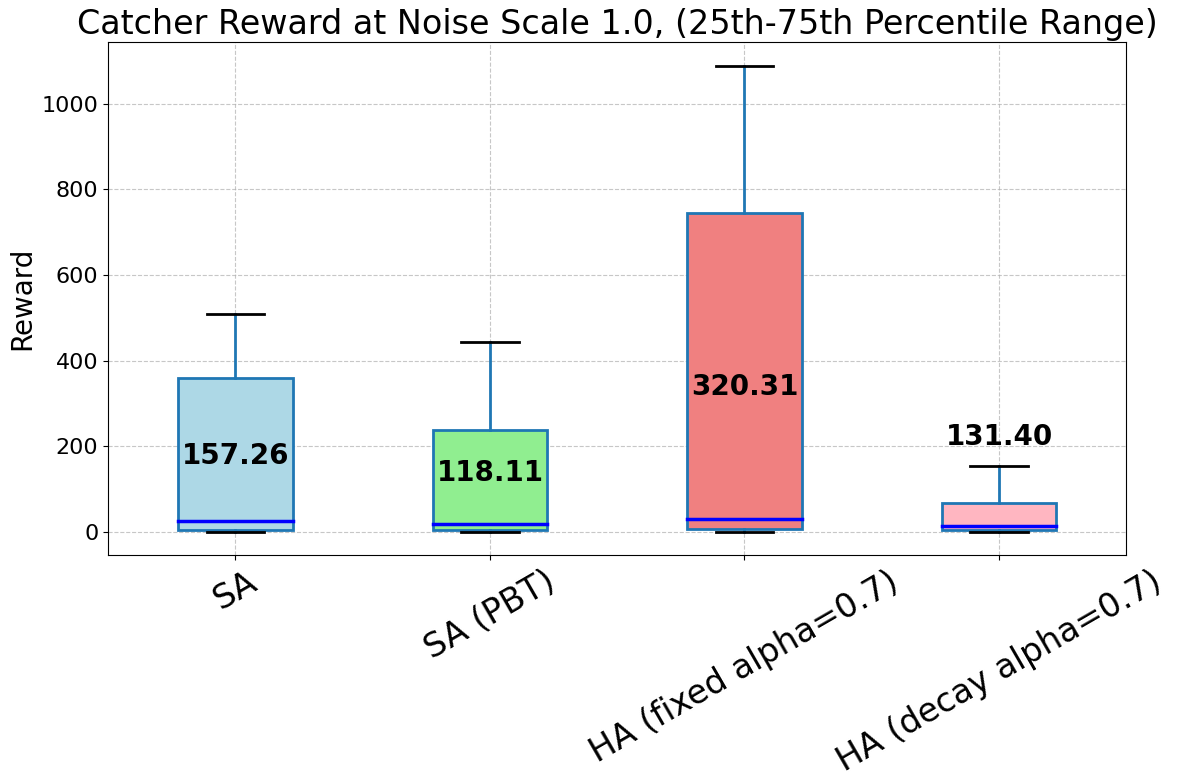}
{\sffamily\footnotesize
\begin{flushleft}
\begin{NiceTabular}{m[c]{0.4cm} m[c]{1.43cm} m[c]{1.43cm} m[c]{1.43cm} m[c]{1.43cm}}
 & SA & SA & HA & HA \\
 & & (PBT) & (Fixed, & (Decay, \\
 & &  & $\alpha$=0.7) & $\alpha$=0.7) \\
\end{NiceTabular}
\end{flushleft}
}
	\end{center}
	\vspace{-0.5em}
	\caption{Box plots of catcher rewards for single-agent and proposed heterogeneous agents. The box represents the interquartile range from the 25th to the 75th percentile, with whiskers extending to the minimum and maximum non-outlier values. The blue line indicates the median reward, while the numbers inside the boxes represent the mean rewards. The proposed HA (Fixed, $\alpha=0.7$) achieves better performance than the other methods.}
        \label{fig:result_sa_ha}
	% \vspace{-1.5em}
\end{figure}

% Catch agent 가 왼쪽으로 던져진 컵을 왼손으로 치고 양손으로 쥐어서 성공적으로 잡았다. 큰 Board 물체를 손으로 막아서 떨어뜨리는데는 성공했지만 손가락으로 움겨쥐는데는 실패했다.

\begin{figure}
    \begin{center}
    \includegraphics[width=\linewidth]{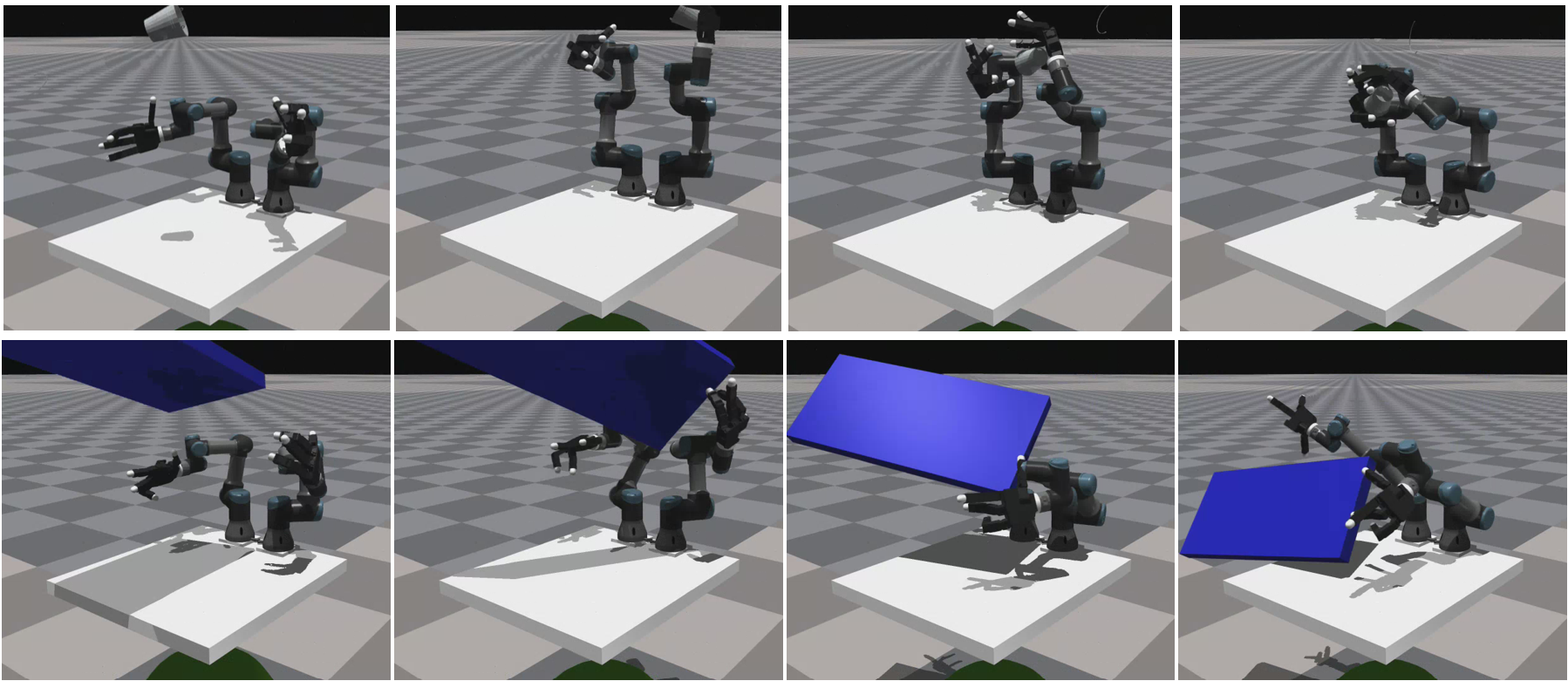}
    \vspace{-2.5em}
    \end{center}
    \caption{Success (top) and failure (bottom) examples. The catch agent hit the cup thrown to the left with its left hand and successfully caught it by gripping it with both hands. The agent successfully blocked the large board object with its hands to prevent it from throwing away but failed to grasp it with its fingers.}
    \label{fig:img_sequence}
    \vspace{-1.5em}
\end{figure}

\subsection{Qualitative Results}
\label{sec:qualitative_results}

Figure \ref{fig:fig4} shows snapshots of the catching for 15 objects, and Figure \ref{fig:img_sequence} shows snapshot sequences of exemplary cases. Please refer to the supplementary video for animated results.

\subsection{Evaluations on Single and Multi-Agent Learning}
\label{sec:evaluation_ma}

We compared the single-agent (SA), with and without population-based training (PBT) \cite{jaderberg2017population}, and the proposed adversarial-cooperative (AC)-HA frameworks for the throw-catch task. The evaluation was based solely on the catcher’s reward, using a baseline uniform random object-throwing scenario. In this environment, the object is respawned near the table’s edge, offset by 20 cm along the x-axis and 30 cm along the z-axis, with uniform random noise. It is thrown toward the catcher with a base linear velocity of [5.0, 2.0, 4.0] m/s and angular velocity of [10.0, 10.0, 10.0] rad/s, both perturbed by uniform noise $\epsilon \sim \mathcal{U}(-0.5, 0.5)$. Success is determined by whether the object remains above a failure threshold, set at half its predefined size relative to the table height; objects falling below trigger a reset. This criterion ensures consistency across object sizes for objective evaluation. The baseline uniform randomization is used for single-agent learning, while all models are evaluated under more challenging, unknown throwing distributions with noise scalings of \{1.0, 1.2, 1.5\} to test robustness.

Figure \ref{fig:result_sa_ha} presents the results as a box plot of reward distributions (25th-75th percentile range) for 1,000 episodes with a noise scale of 1.0. Despite being the baseline learning environment for the single-agent model, the proposed AC-HA significantly outperforms the others. Even though we have filtered out many high-reward samples by only showing the 75th percentile range, the AC-HA model still exhibits relatively higher maximum reward episodes. This suggests that the thrower agent, under the proposed adversarial-cooperative reward function, has learned to cooperate effectively during the training process. The HA model used in the evaluation fixes the adversarial-cooperative weighting parameter $\alpha$ at 0.7. 

To explore the impact of varying $\alpha$, we trained an alternative HA model over 10,000 epochs, where $\alpha$ was linearly decayed from 1.0 to 0.7 (denoted as HA (decay $\alpha=0.7$) in the figure) under the same conditions. The results showed significantly worse performance than the fixed-$\alpha$ model, even underperforming the single-agent model. This suggests that changing $\alpha$ made it difficult for the thrower to make consistent adversarial-cooperative decisions, leading to unstable learning. This instability is evident from lower maximum rewards, particularly in early training when $\alpha=1.0$, where the thrower initially benefited from cooperative behavior but struggled as training shifted toward a more adversarial setting. We interpret this as the thrower agent being unable to adapt effectively as the weighting criterion changed. Ultimately, our experiments demonstrate that dynamically adjusting $\alpha$ in multi-agent learning introduces instability, as reflected in performance degradation.

% To explore the impact of varying $\alpha$, we trained an alternative HA model over 10,000 epochs where $\alpha$ was linearly decayed from 1.0 to 0.7 (denoted as HA (decay $\alpha=0.7$) in the figure) under the same conditions. The results showed significantly worse performance compared to the fixed-$\alpha$ model, even underperforming the single-agent model. This indicates that the changing $\alpha$ made it difficult for the thrower to make consistent decisions from an adversarial-cooperative perspective, leading to unstable learning outcomes. This instability is evident from the lower maximum rewards, particularly in the early training stages when $\alpha=1.0$, where the thrower initially benefited from purely cooperative behavior but struggled as the training progressed toward a more adversarial setting. We interpret this as the thrower agent being unable to adapt its policy effectively as the weighting criterion shifted. Ultimately, our experiments reveal that dynamically adjusting $\alpha$ in multi-agent learning introduces instability, as demonstrated by the performance degradation.

% We performed evaluations using uniform random throws, varying the noise scale of the throw agent’s actions across {1.0, 1.2, 1.5}. The performance of single-agent, where only catcher is learned, and multi-agent configurations ($\alpha=0.9$ for the reward balance) is compared. The results are analyzed by adjusting the noise scale during uniform random throw tests, demonstrating the effectiveness of the adversarial training approach.

\begin{figure}
	\begin{center}
		\includegraphics[width=\linewidth, trim=0cm 7.4cm 0cm 0cm, clip]{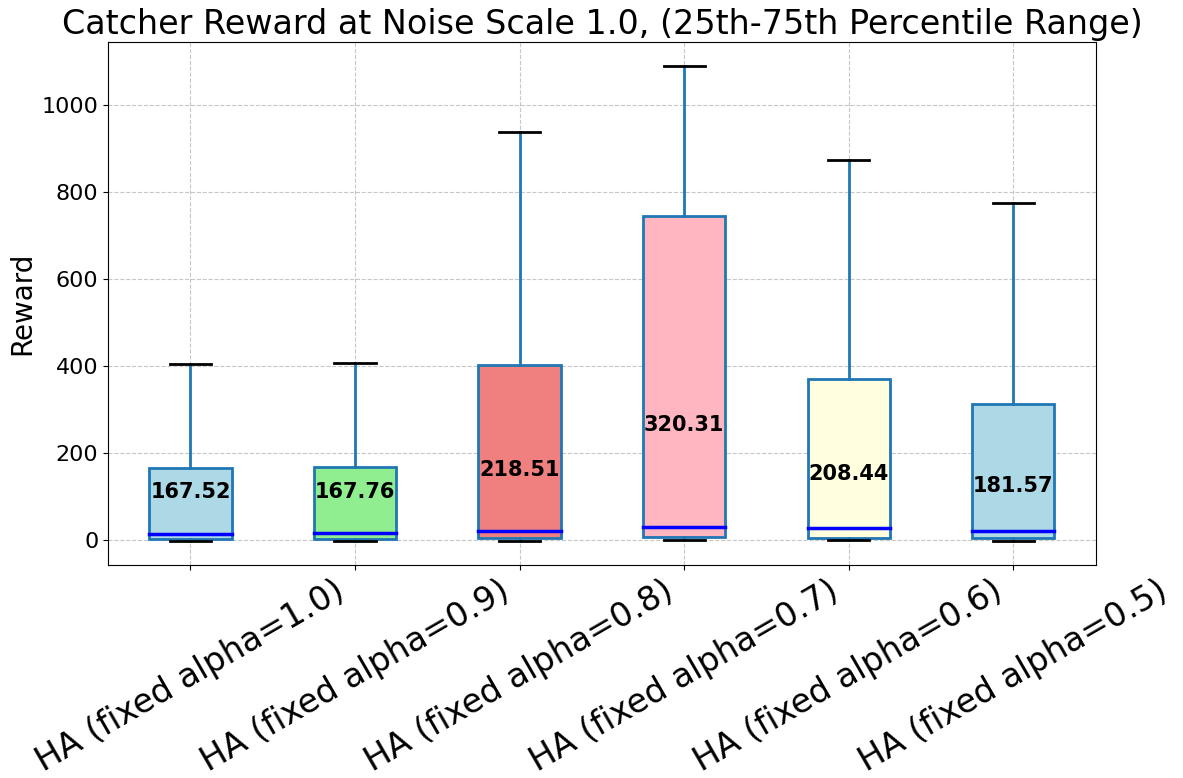}
	\end{center}
{\sffamily\footnotesize
\begin{flushleft}
\begin{NiceTabular}{m[c]{0.34cm} m[c]{0.78cm} m[c]{0.78cm} m[c]{0.78cm} m[c]{0.78cm} m[c]{0.78cm} m[c]{0.78cm}}
 & HA & HA & HA & HA & HA & HA \\
 & (Fixed, & (Fixed, & (Fixed, & (Fixed, & (Fixed, & (Fixed, \\
 & $\alpha$=1.0) & $\alpha$=0.9) & $\alpha$=0.8) & $\alpha$=0.7) & $\alpha$=0.6) & $\alpha$=0.5) \\
\end{NiceTabular}
\end{flushleft}
}
\vspace{-0.5em}
	\caption{Box plots of catcher rewards with different $\alpha$ values. The results show that $\alpha$=0.7 yields the best performance.}
        \label{fig:results_alpha}
	\vspace{-1.5em}
\end{figure}

\subsection{Ablations}

% $\alpha$에 따른 성능 차이를 비교하기 위하여, $\alpha=1.0$부터 $\alpha=0.5$에 대한 성능을 비교하였다. Figure 6.은 HA의 fixed $\alpha$값 변화에 따른 성능 변화를 25-75 percentile 범위의 box plot으로 나타낸 것이다. Catcher에 완전히 cooperative한 $\alpha=1.0$보다는 절반의 적대성(adversarial)을 유도하는 $\alpha=0.5$ 가 평균적으로 보다 높은 성능을 보인다. 또한 적당한 수준의 적대성, 즉 20~40%의 적대성을 보이는 것이 완전히 협조적인 것 보다는 높은 성능을 보이는데, 이를 통해 적대성이 catcher의 스킬 향상에 도움이 됨을 실험적으로 검증하였다. 반면, 지나치게 적대성을 높이는 경우에는 오히려 성능을 감소시키는 경향을 보이는데, 이와 같은 매커니즘은 인간이 어떤 지식이나 기술을 습득할 때, 적절한 수준의 어려움을부여하는 것이 장기적으로는 기술 향상에 도움이 되는 것과 유사한 맥락이다. 

% 우리는 catcher 에이전트가 학습 과정에서 경험해보지 못했던 보다 도전적인 과제를 직면했을 때, single-agent policies 뿐만 아니라 $\alpha$에 따른 multi-agent policy가 어떠한 성능 변화를 보이는지 확인하기 위하여, 기존 base uniform random object throwing에 [1.0, 1.2, 1.5]의 noise scale을 곱하여 보다 어려운 던지기 환경을 구성하고 각 정책별로 성능을 비교분석하였다. 실험 결과, [1.2, 1.5]의 보다 도전적인 환경에서도 multi-agent가 single-agent보다 전반적으로 높은 성능을 보였으며, HA $\alpha=0.7$가 전 과정에 걸쳐 가장 높은 성능을 유지했다. $\text{noise\_scale}=1.5$와 같이 난이도가 올라갈 수록 $\alpha$값에 따른 성능 차이는 줄어드는 경향을 보였으나, 전반적으로 single-agent보다는 높은 성능을 유지했다. 이는 catcher가 학습과정에서 thrower의 적대성에 의해 도전적인 환경을 경험했기 때문으로 해석된다. 본 실험을 통해 우리는 적절한 수준의 적대성과 동시에 협조성을 보유한 opposite agent가 learner의 학습에 효과적일 수 있는 실험적 근거를 마련하였다. 

To compare the performance differences based on the value of $\alpha$, we evaluated performance across $\alpha$ values from 1.0 to 0.5. Figure \ref{fig:results_alpha} shows the performance variation in the HA framework with different fixed $\alpha$ values, represented as box plots covering the 25th to 75th percentile range. The results demonstrate that $\alpha=0.5$, which induces partial adversarial behavior, outperforms the fully cooperative $\alpha=1.0$ in terms of average performance. Furthermore, a moderate level of adversariality, around 20-40\%, results in higher performance compared to fully cooperative behavior. This experimentally validates that some degree of adversariality contributes to the improvement of the catcher’s skill. On the other hand, excessively increasing adversariality tends to degrade performance. This mechanism is analogous to human learning, where providing an appropriate level of challenge enhances long-term skill acquisition \cite{vygotsky1978mind, bjork1994memory}.

\begin{figure}
	\begin{center}
		\includegraphics[width=\linewidth]{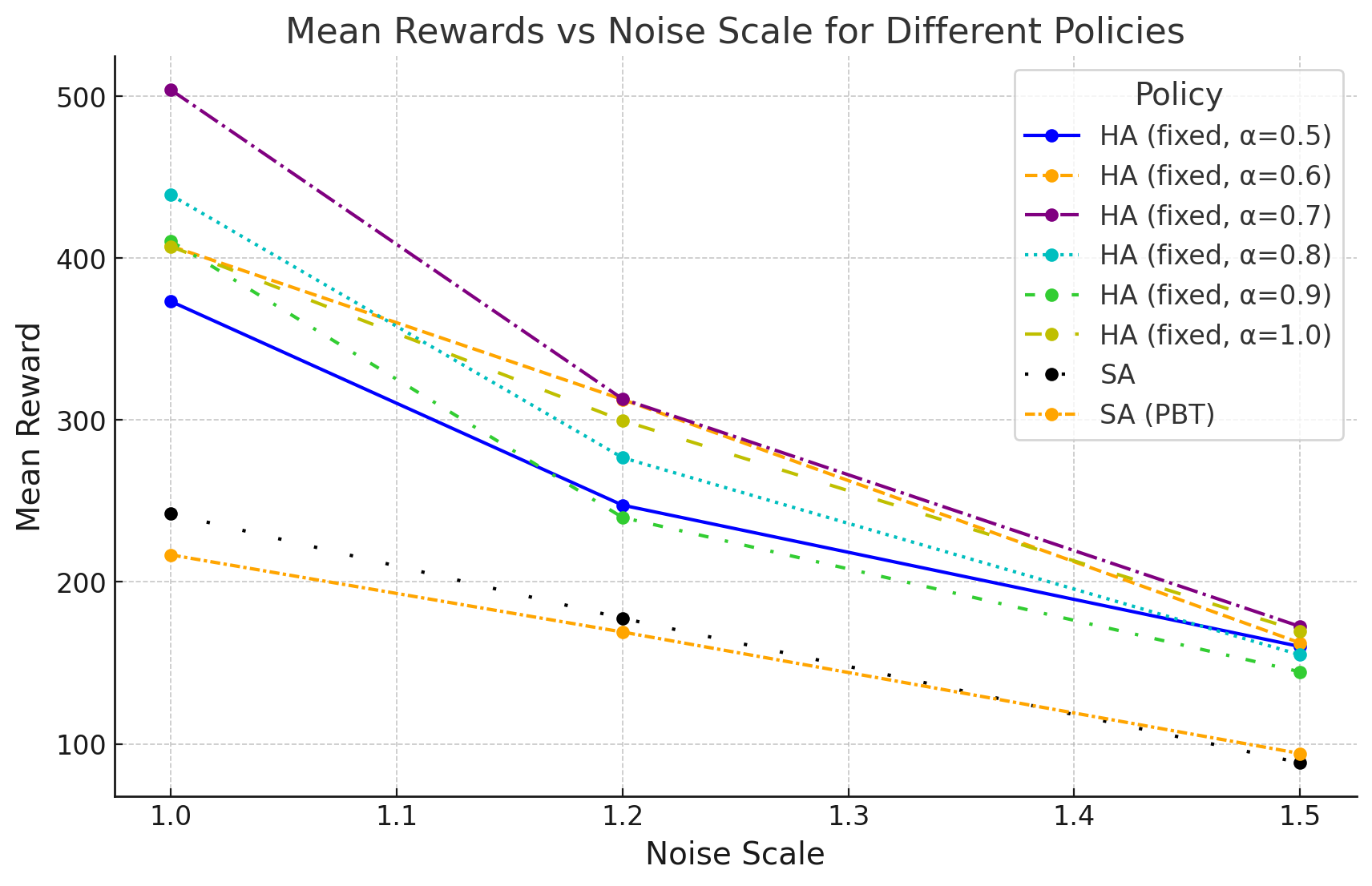}
	\end{center}
	\vspace{-1.5em}
	\caption{\text{HA (fixed, $\alpha=0.7$)} achieved the highest performance, indicating that setting a moderate adversarial weight (e.g., 0.7 or 0.8) generally improves the performance of learning catch skills.}
        \label{fig:noise_scale}
	\vspace{-1.5em}
\end{figure}

To further investigate how the multi-agent policy with varying $\alpha$ values performs, especially when the catcher agent encounters more challenging tasks not experienced during training, we introduced higher difficulty environments by applying noise scales of \{1.0, 1.2, 1.5\} to the base uniform random object throwing setup. We then compared the performance of each policy. The experimental results show that even in the more challenging environments with noise scales of \{1.2, 1.5\}, the multi-agent approach consistently outperforms the single-agent approach (see Figure \ref{fig:noise_scale}). Notably, the HA framework with $\alpha=0.7$ achieved the best performance overall. Although the performance gap between different $\alpha$ values decreased as the difficulty increased (e.g., with noise scale=1.5), the multi-agent models maintained overall higher performance than the single-agent model. This suggests that the challenging environments encountered by the catcher during training, due to the thrower’s adversarial behavior, were beneficial for its skill development. Our experiments provide empirical evidence that an opposite agent, balancing both adversariality and cooperation, can effectively enhance the learner's performance.

% Ablation studies are conducted by varying $\alpha$ parameter to explore its impact on system performance. We analyze how changes in $\alpha$ affect the catch agent’s ability to adapt to increased noise scales in the throw agent’s behavior, highlighting the significance of the adversarial reward mechanism.

\subsection{Object-Specific Catch Skill}

\begin{figure}
	\begin{center}
		\includegraphics[width=\linewidth]{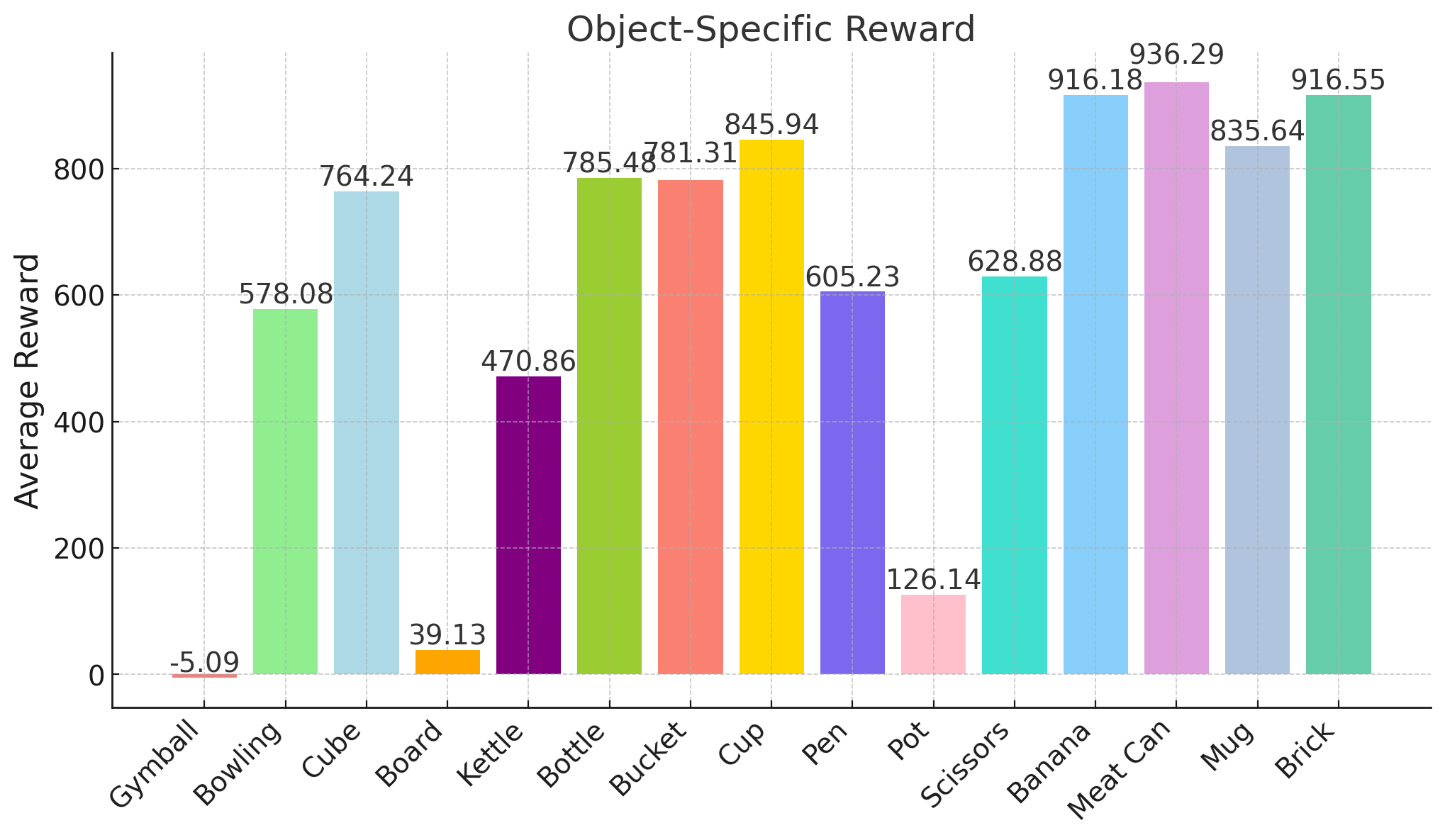}
	\end{center}
	\vspace{-1.5em}
	\caption{Average rewards for each object over 1000 episodes. Smaller objects like `Meat Can,' `Brick,' and `Banana' yield higher rewards, while larger objects such as `Gymball,' `Board,' and `Pot' result in lower rewards}
        \label{fig:object_specifc_rew}
	\vspace{-1.5em}
\end{figure}

The variation in object shapes leads to differences in catching strategies for each object. We conducted a per-object evaluation of catch performance and quantitatively analyzed these differences (see Figure \ref{fig:object_specifc_rew}). Our results indicate that smaller objects, such as cubes and bottles, typically yield higher rewards, whereas larger objects like gymballs, boards, and pots exhibit noticeably lower performance. This disparity is more likely a result of dataset imbalance rather than the inherent difficulty of learning catch skills for larger objects. Notably, when training was restricted to the largest objects—gymball and board—the agents displayed competent catching abilities (see \href{https://youtu.be/LbzgZB7IQuY}{Video}). This implies that catch strategies are highly dependent on the object's size and shape. Addressing the challenges posed by dataset imbalance and strategy variation across different objects will be a subject of future research.

% 물체의 shape이 서로 다르기 때문에 각 물체별로 catch 전략이 달라질 수 있다. 우리는 물체별로 catch skill을 평가 하여 그 차이를 정량적으로 비교했다. 물체별 실험 결과, cube, bottle과 같이 작은 물체들은 대체적으로 높은 보상을 보이지만, gymball, board, pot과 같은 상대적으로 크기가 큰 물건은 매우 낮은 성능을 보인다. 이는 해당 물체에 대한 catch skill을 학습하기 어렵다기 보다는 전체 데이터셋의 불균형으로 인하여 야기된 결과라고 할 수 있다. 그 근거로써, 크기가 가장 큰 gymball과 board object만을 대상으로 학습을 했을 때에는 비교적 잘 잡는 모습을 보였기 때문이다. 즉 object 크기나 모양에 따라 catch 전략이 상이한 분포를 보이기 때문으로 해석된다. 다양한 object에 대한 데이터 불균형 및 서로 다른 전략 문제는 추후 연구 주제중 하나로 남겨놓는다. 

\section{Conclusion \& Limitations}

In this paper, we presented an adversarial cooperation system for learning bimanual catch skills using Heterogeneous-Agent Reinforcement Learning (HARL). Our approach introduces an adversarial-cooperative reward structure that enhances dexterous catching under dynamic conditions. Simulations showed significant improvements in bimanual coordination and adaptability over single-agent methods.

A key advantage of our method is that the thrower agent generates an adaptive curriculum through adversarial-cooperative interactions. Unlike handcrafted curricula, our approach dynamically adjusts challenge levels based on the catcher’s evolving skill, ensuring continuous learning efficiency and robust generalization. Results confirm that the learned curriculum optimizes throwing difficulty, leading to superior catching performance over static policies.

Despite these advancements, limitations remain. All experiments were conducted in simulation, requiring adaptation for real-world deployment. The object set, though diverse, was limited to 15 items, and future work should assess broader generalization. Additionally, our framework assumes stationary agents, which may limit performance in complex scenarios. While our study focused on a throwing and catching task, the underlying adversarial-cooperative learning framework could be extended to other domains, such as collaborative assembly, dexterous manipulation, or multi-agent coordination. Exploring such generalizations remains an exciting research direction.

Future work will address these limitations by deploying the framework on real robots and extending it to dynamic environments, including mobile manipulators and humanoids, broadening its impact on robotic learning and interaction.

\section*{ACKNOWLEDGMENT}

% This work was supported by Electronics and Telecommunications Research Institute (ETRI) grant funded by the  Korean government foundation. [24ZB1200, Research of  Human-centered autonomous intelligence system original  technology, 70\%] and the Institute of Information \& communications Technology Planning \& Evaluation(IITP) grant funded by the Korea government(MSIT) [No. RS-2024-00336738, Development of Complex Task Planning Technologies for Autonomous Agents, 30\%]

This work was partly supported by Electronics and Telecommunications Research Institute (ETRI) [24ZB1200, Research of Human-centered Autonomous Intelligence System Original Technology, 70\%] and the Institute of Information \& Communications Technology Planning \& Evaluation (IITP) grant funded by the Korea government (MSIT) [RS-2024-00336738, Development of Complex Task Planning Technologies for Autonomous Agents, 30\%].

% This work was supported by Electronics and Telecommunications Research Institute (ETRI) grant funded by the Korean government foundation. [24ZB1200, Research of Human-centered autonomous intelligence system original technology, 70\%] and [24BB1700, Development of Multimodal Interactive Robot Artificial Intelligence Technology, 30\%]  

\bibliographystyle{IEEEtran}
\bibliography{refs.bib}

\end{document}